\documentclass[10pt,twocolumn,letterpaper]{article}

\usepackage{cvpr}              

\usepackage{graphicx}
\usepackage{amsmath}
\usepackage{amssymb}
\usepackage{booktabs}

\usepackage[pagebackref,breaklinks,colorlinks]{hyperref}

\usepackage{multirow}
\usepackage[capitalize]{cleveref}
\crefname{section}{Sec.}{Secs.}
\Crefname{section}{Section}{Sections}
\Crefname{table}{Table}{Tables}
\crefname{table}{Tab.}{Tabs.}

\begin{document}

\title{Adaptive Graph-Based Feature Normalization for Facial Expression Recognition}

\author{Yangtao Du\\
Fudan University\\
{\tt\small 19210860029@fudan.edu.cn}
\and
Qingqing Wang*\\
Shanghai Jiao Tong Univeristy\\
{\tt\small qqwang0723@sjtu.edu.cn}
\and
Yujie Xiong\\
Shanghai University of Engineering Science\\
{\tt\small  xiong@sues.edu.cn}
}
\maketitle

\begin{abstract}
Facial Expression Recognition (FER) suffers from data uncertainties caused by ambiguous facial images and annotators' subjectiveness, resulting in excursive semantic and feature covariate shifting problem. Existing works usually correct mislabeled data by estimating noise distribution, or guide network training with knowledge learned from clean data, neglecting the associative relations of expressions. In this work, we propose an \textbf{A}daptive \textbf{G}raph-based \textbf{F}eature \textbf{N}ormalization (AGFN) method to protect FER models from data uncertainties by normalizing feature distributions with the association of expressions. Specifically, we propose a Poisson graph generator to adaptively construct topological graphs for samples in each mini-batches via a sampling process, and correspondingly design a coordinate descent strategy to optimize proposed network. 
Our method outperforms state-of-the-art works with accuracies of 91.84\% and 91.11\% on the benchmark datasets FERPlus and RAF-DB, respectively, and when the percentage of mislabeled data increases (e.g., to 20\%), our network surpasses existing works significantly by 3.38\% and 4.52\%.
\end{abstract}

\section{Introduction}

Facial expression is a natural signal to convey emotions and intentions of human beings. Therefore, facial expression recognition (FER) is essential for machines to understand human behaviors and interact with humans. Though great efforts have been made in last decades and promising progress has been achieved, FER still suffers from data uncertainty problem, i.e., data is frequently mislabeled because of the subjectiveness of annotators and the ambiguities of facial images. Existing works on FER rarely focus on this problem, except IPA2LT~\cite{r03} and Self-Cure Network (SCN)~\cite{r01}, which suppress data uncertainties by discovering the latent truth from inconsistent pseudo labels or relabeling mislabeled data by weighting and ranking samples in a mini-batch. Though learning with noisy labels has been studied extensively in the community of computer vision, existing works mainly focus on correcting mislabeled data by estimating label quality and noise distribution, or guiding network training with knowledge learned from clean data. However, as pointed in~\cite{r02}, when people encounter a vague facial image with fuzzy expression, they often associate it with other images sharing similar expressions, instead of staring at its parts for research, which is called human adaptively associative learning process. In other words, humans tend to make associative comparisons when discerning subtle expressions, while current works neglects the associative relations of facial images. 

\begin{figure}[tp]
\centering
\includegraphics[width=8cm]{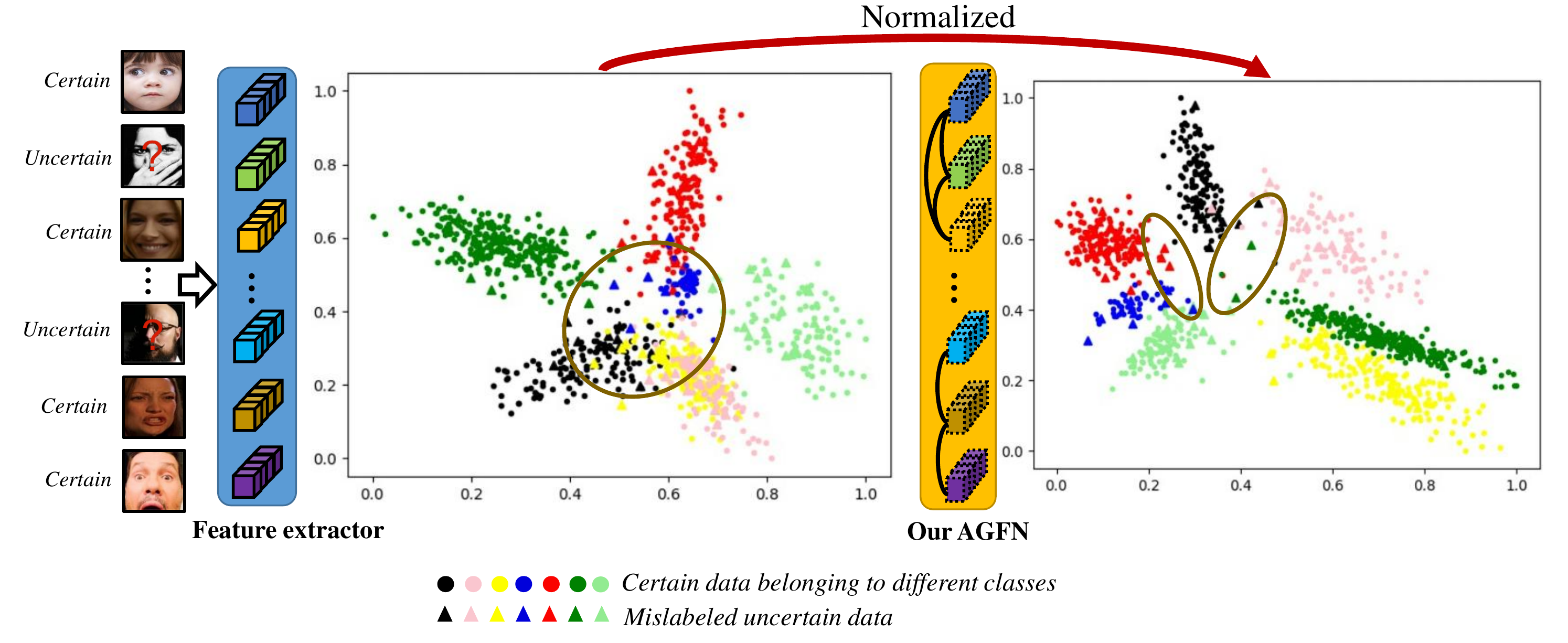}
\caption{The excursive semantic and featuer covariate shifting problem caused by uncertain data. After being normalized by our AGFN, feature distribution is with much clearer boundaries although mislabeled samples are still there.}
\label{fig_1}
\end{figure}

Moreover, with the growth in scale of training samples gathered from internet, data uncertainties have been introducing great challenges to FER by leading to excursive semantic and feature covariate shifting, as shown in Figure~\ref{fig_1}, where distributions of individual classes are with serious overlaps because of mislabeled data. Therefore, in this work, we propose an efficient normalization method called Adaptive Graph-based Feature Normalization (AGFN) to tackle the data uncertainty problem by normalizing feature distributions with the association of expressions. As shown in Figure~\ref{fig_1}, with the assistance of proposed AGFN, individual classes can be split with much clearer boundaries though mislabeled data still exists.  

Specifically, given feature maps extracted from facial images, AGFN firstly projects them into emotional feature vectors. Then, under the assumption that the probability of sample connections satisfies Poisson distribution and corresponding parameters are closely related to samples' similarities, a Poisson graph generator is designed to adaptively construct topological graphs for samples in each mini-batches. Afterwards, Graph Convolutional Network (GCN) is exploited to convey the semantic information of associated samples because expressions present in facial images can be reflected by other images sharing similar features in a proper feature space. In addition, since the calculation of adjacent matrices used for graph generation involves a sampling process, parameters of our network can not be optimized by the widely used gradient decent method. Therefore, we design a coordinate decent strategy to jointly optimize parameters of neural networks and the sampling process.   
According to our experiments, when equipping a naive FER model with proposed AGFN, its recognition performance can be improved by large margins from 85.37\% to 91.84\% and from 85.89\% to 91.11\% on the benchmark datasets FER2013plus and RAF-DB, implying the importance of tackling the uncertainty problem in FER as well as the effectiveness of proposed AGFN. Moreover, we also conduct experiments on synthetic datasets where a large portion of samples are mislabeled, finding that our AGFN-equipped network surpasses existing state-of-the-art works significantly by 3.38\% and 4.52\% on data with serious uncertainties. 
Our contributions are as follows:

(1) We propose to utilize the associative relations of expressions to tackle the excursive semantic and feature covariate shifting problem caused by data uncertainties, and propose an effective normalization method named AGFN to elevate the performance of FER models; 

(2) We design a Poisson graph generator to adaptively construct topological graphs for samples in each mini-batches with a sampling process, and utilize GCN to normalize feature distributions. Moreover, we design a coordinate decent strategy to jointly optimize parameters involved in neural networks and sampling process;  

(3) We conduct extensive experiments to show the superiority of proposed AGFN, and demonstrate that when a large portion (e.g. 20\%) of samples are mislabeled, our AGFN-equipped network shows much better robustness and effectiveness.

\section{Related works}

\subsection{Facial Expression Recognition}

Feature extraction and expression classification are the two basic modules of typical FER pipeline, and in past years, most of related works have focused on designing more effective and robust feature extractors. For example, works~\cite{r05}~\cite{r06}~\cite{r07}~\cite{r08} explored various deep neural networks to extract more powerful features, including VGG network, Inception network, Residual network and Capsule network etc. Dinesh et al.~\cite{r09} pointed out that the widely used convolutional layers and average pooling layers only captured first-order statistics,  so they proposed to extract the second order statistic features with covariance pooling. In practice, pose variations, occlusions and uneven illuminations always resulted in low quality facial images, on which FER models usually failed to extract discriminative features. Therefore, Wang et al.~\cite{r10} designed a regional attention network to improve the performance of FER. Inspired by the psychological theory that expressions could be decomposed into multiple facial action units, Liu et al.~\cite{r11} constructed a deep network called AU-inspired Deep Networks (AUDN) to combine the informative local appearance variation and high level representation. Generally, objective functions of FER networks considered each samples independently, while Zhao et al.~\cite{r12} addressed this issue from another perspective. They designed a peak-piloted deep network (PPDN) to supervise the intermediate feature responses for samples with non-peak expression. Recently, Transformer~\cite{r26} showed its power in the field of Neural Language Processing (NLP) and Computer Vision (CV). Therefore, Ma et al.~\cite{r25} and Huang et al.~\cite{r22} also introduced Visual Transformer to FER and achieved promising performance.

\subsection{Uncertainties in Facial Expression Recognition}

Uncertainties result in mislabeled data, which seriously affects the performance of FER models. Though learning with noisy labels has attracted extensive attentions in the community of CV, it is rarely studied in the task of FER. On the other hand, existing works on general CV tasks mainly focus on addressing this issue by pre-training networks on weak data and then fine-tuning them with true labels~\cite{r14}, guiding the training of networks with knowledge learned from clean data~\cite{r13}, or relabeling mislabeled data together with learning powerful representations and classifiers~\cite{r16}. For example, to alleviate the harm from noisy data, Dehghani et al.~\cite{r13} updated the gradients of a target network under the guidance of a second confidence network that trained with a small set of clean data. Li et al.~\cite{r15} designed a unified distillation framework, where the distillation process was guided with a knowledge graph, to `hedge the risk' of learning from noisy labels. Apparently, above works tried to estimate label quality or noisy distribution with a small set of clean data, while works without utilizing clean data usually introduced additional constrains or distributions on noisy data~\cite{r17}~\cite{r18}. For example, Mnih et al.~\cite{r17} proposed more robust loss functions to deal with omission noise and registration noise on aerial image datasets. To estimate correct labels, Goldberger et al.~\cite{r18} viewed the correct label as latent random variable and modeled the noise processes by a communication channel with unknown parameters, which were optimized with EM algorithm. For the task of FER, Zeng et al.~\cite{r03} was the first to improve FER performance by addressing the data uncertainty issue. They assigned more than one labels to each samples and discovered the latent truth from the inconsistent pseudo labels with an end-to-end LTNet. Subsequently, Wang et al.~\cite{r01} suppressed data uncertainties by weighting and ranking samples in a mini-batch with a self-attention mechanism, followed by modifying samples' labels in the lowest-ranked group. 

In summary, previous works have explored how to discover the truth of mislabeled data and prevent networks from the harm of noisy data, neglecting the associative relations of expressions. In contrast, our AGFN protects FER models from data uncertainties by tackling the excursive semantic and feature covariate shifting problem with considering the association of expressions.    

\begin{figure*}[tp]
\begin{center}
\includegraphics[width=15cm]{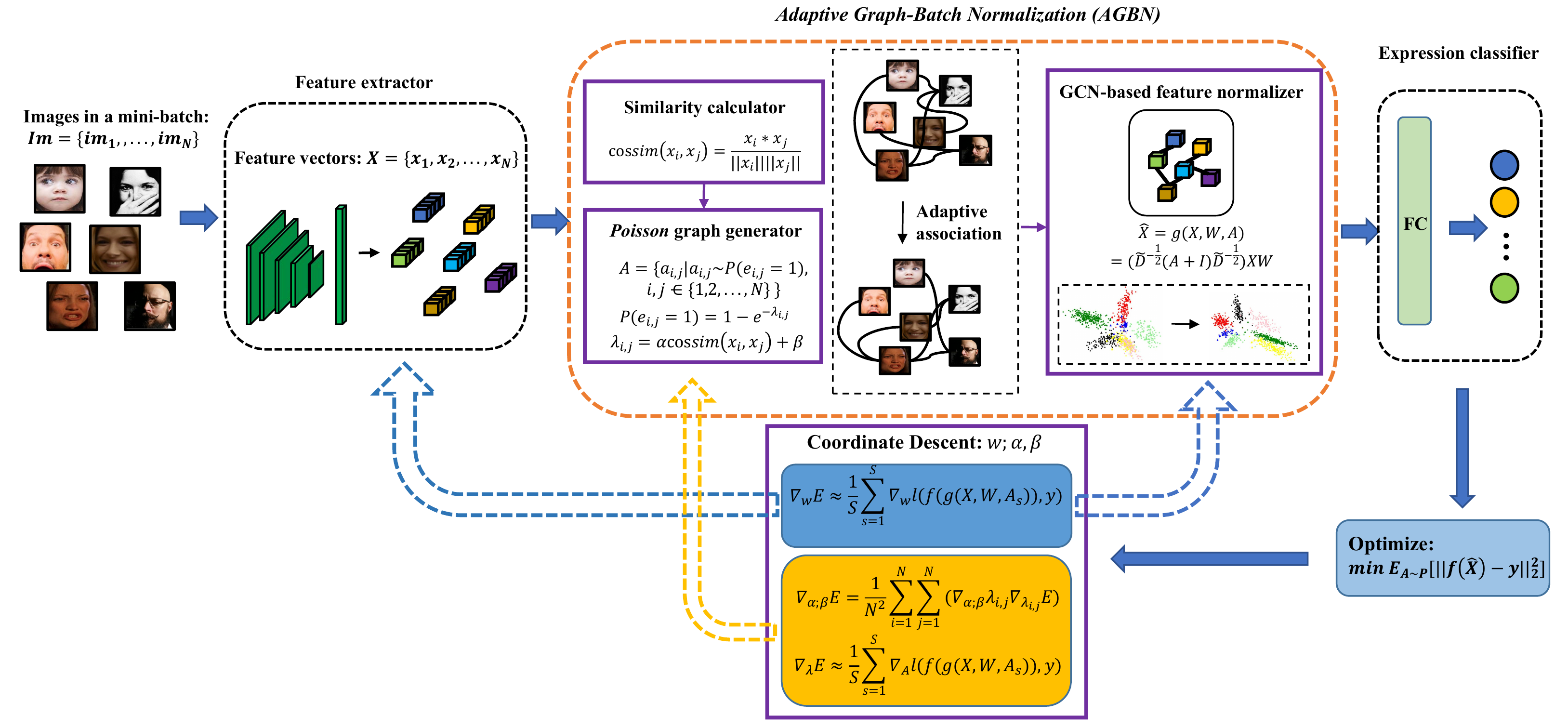}
\end{center}
   \caption{Architecture of proposed AGFN-equipped network. The AGFN module is composed of similarity calculator, Poisson Graph generator and GCN-based feature normalizer, and can be conveniently inserted into FER models between their feature extractors and expression classifiers. }
\label{fig:short}
\end{figure*}

\section{Method}

Facial images containing similar subtle expressions are most likely to share the same labels and according to human associative learning mechanism~\cite{r02}, humans tend to correlate objects with similar abstract features. Therefore, exchanging semantic information among samples with high similarity can help to normalize features of individual samples, leading to improvement of FER performance. Toward this end, we design a feature normalization method called Adaptive Graph-based Feature Normalization (AGFN). Given a baseline model that composed of a feature extractor and an expression classifier, which are basic components used to extract feature maps from facial images and distinguish corresponding expressions, our AGFN can be conveniently inserted into it as shown in Figure~\ref{fig:short}. Specifically, AGFN exploits a novel graph generator to dynamically and adaptively construct topological graphs for samples in each mini-batches according to their similarities. In this generator, adjacent matrices of topological graphs are determined by a sampling process. Then, GCN is used to transfer semantic information among associated samples. Since gradient calculation rules of parameters from neural networks and above sampling process are different, traditional gradient decent method is not applicable anymore. Therefore, we propose a coordinate decent strategy to optimize our network in an end-to-end way. 

\subsection{Poisson Graph Generator}

Traditional graph-based methods usually connect samples with high similarities with the widely used threshold-based staircase function. However, samples with very similar features may not belong to the same classes and high similarities only imply high probabilities of sharing the same class labels. To address this issue, we propose to model the relations between feature similarities and sample connection probabilities with Poisson distribution. Poisson distribution is used to describe times that an event happens in a unit time, as shown in Eq.~\ref{e0}, where $k$ and $\lambda$ denote times and average times that an event happens per unit time. In human associative learning, two samples are usually compared for multiple times to confirm whether they belong to the same classes, and different regions of interest are looked every time. Therefore, we assume that probabilities of sample connections satisfy Poisson distribution and corresponding parameters are closely related to samples' similarities. Subsequently, a novel Poisson graph generator is proposed to calculate the adjacent matrices of topological graphs with a sampling process.
\begin{equation}
Po(X=k;\lambda)=\frac{\lambda ^k}{k!}e^{-\lambda},k=0,1,...K
\label{e0}
\end{equation}

Given input images $Im=\{im_1, im_2,..., im_N\}$ in a mini-batch, the FCN-based feature extractor outputs corresponding feature maps $M=\{m_1, m_2,..., m_N\}$, which is further projected into emotional feature vectors $X=\{x_1, x_2,..., x_N\}$ by a Multi-Layer Perceptron (MLP). Then, similarity $cossim(x_i, x_j)$ between sample $im_i$ and $im_j$ can be calculated with the cosine similarity coefficient formulated in Eq.~\ref{e1}:
  
\begin{equation}
    cossim(x_i, x_j)=\frac{x_i*x_j}{||x_i||||x_j||}
\label{e1}
\end{equation}

Intuitively, we associate an object to different ones for multiple times to capture more detailed information. Therefore, for better robustness, the construction of topological graphs should follow a stochastic mechanism rather than the widely used threshold-based staircase function so that different contrastive objects can be seen in different iterations. Hence, we model the connection probability $p(e_{i,j}=1)$ of sample $im_i$ and sample $im_j$ with Poisson distribution, as shown in Eq.~\ref{e2}, and the Poisson parameter ${\lambda_{i,j}}$ is computed with the similarity $cossim(x_i, x_j)$ via a linear function described in Eq.~\ref{e3}. Here, parameters ${\alpha}$ and ${\beta}$ are introduced to scale the probability distribution, and will be learned during the training procedure.

\begin{equation}
\begin{aligned}
 P(e_{i,j}=1) & =1-Po(0;\lambda_{i,j}) \\
	             & = 1-\frac{e^{-\lambda_{i,j}}\lambda_{i,j}^0}{0!}=1-e^{-\lambda_{i,j}}
	\end{aligned}
\label{e2}
\end{equation} 

\begin{equation}
      \lambda_{i,j}=\alpha cossim(x_i, x_j) + \beta
\label{e3}
\end{equation}

Afterwards, we sample the adjacent matrix ${A}$ according to $A\sim P$ (see Eq.~\ref{e4}) for samples in current mini-batch. Expectations of the sampling process will be optimized as introduced in Section~\ref{pipeline}.
\begin{equation}
\begin{aligned}
 A= \{a_{i,j}|a_{i,j}\sim P(e_{i,j}=1), i,j\in\{1,2,...,N\}\}
	\end{aligned}
\label{e4}
\end{equation} 

\subsection{Feature Normalization with GCN}

To alleviate the excursive semantic and feature covariate shifting problem, we imitate human associative learning procedure by conveying semantic information among associated samples with GCN~\cite{r34}, which is built upon samples' topological graphs generated by our Poisson graph generator. The employed GCN is in the second order Chebyshev expansion, as formulated in Eq.~\ref{e5}, where ${A}$ is the adjacent matrix obtained by above sampling strategy, ${W}$ is trainable parameters, $I$ is an identity matrix, $\widetilde{D}$ is a diagonal matrix, ${X}$ denotes samples' emotional feature vectors and $\hat{X}$ represents the expected normalized features.
Here, $\widetilde{D}^{\frac{-1}{2}}$ is used to weight information from associated samples. For $i$-th sample, more associated samples result in greater value of $\widetilde{D}_{ii}$, which means less information from associated samples will be passed to current sample.
\begin{equation}
\begin{aligned}
& \hat{X}=g(X, W, A)= (\widetilde{D}^{\frac{-1}{2}}(A+I)\widetilde{D}^{\frac{-1}{2}})XW \\
		& {\widetilde{D}_{ii}=1+\sum_j{A_{i,j}}}
\end{aligned}
\label{e5} 
\end{equation}

\subsection{Optimization with Coordinate Decent}\label{pipeline}

Suppose our loss function is defined as Eq.~\ref{e6}, where $f(.)$ denotes the expression classifier, then our final goal is twofold: 1) optimizing parameters $W$ involved in neural networks, including feature extractor, GCN and expression classifier; and 2) learning parameters $\alpha$ and $\beta$ used to find the best adjacent matrix ${A \in \mathcal{H}_N}$ in Eq.~\ref{e2} and~\ref{e3}. Here, ${\mathcal{H}_N}$ is the convex hull of the set of all possible adjacency matrices under the Poisson distribution. Furthermore, the objective of our network is to minimize the expectation formulated in Eq.~\ref{e7}.
\begin{equation}
\ell(f(\hat{X}), y)=||f(\hat{X})-y||_2^2
\label{e6} 
\end{equation}

\begin{equation}
  J=\mathop{min}\limits_{W, \alpha,\beta} E_{A\sim P}[\ell(f(\hat{X}), y)]
\label{e7} 
\end{equation}

Since the gradient calculation rule of $W$ is different from that of $\alpha$ and $\beta$, traditional gradient decent strategy is not applicable in our optimization procedure. Therefore, under the assumption that parameters to be optimized are independent from each other, we design a coordinate descent strategy to optimize our network in an end-to-end way. Concretely,  we update $W$ with the tractable approximate learning dynamics shown in Eq.~\ref{e8}, and obtain the approximate gradient $\nabla_WE$ with Eq.~\ref{e9}, where $P(A)$ is the probability of sampling $A$ from distribution $P$ with Eq.~\ref{e4}, ${S}$ represents the pre-defined sampling times and $A_s$ denotes the result of the $s$-th sampling. 
\begin{equation}
\hat{W} = W - \gamma_1\nabla_WE
	\label{e8} 
\end{equation}

\begin{equation}
	\begin{split}
	 \nabla_WE &= \nabla_WE_{A\sim P}[\ell(f(\hat{X}), y)]\\
            &=\sum P(A)\nabla_W \ell(f(g(X, W, A)), y) \\
            &\approx \frac{1}{S}\sum_{s=1}^{S}\nabla_W \ell(f(g(X, W, A_s)), y)
	\end{split}
	\label{e9} 
\end{equation}

 On the other hand, we update $\alpha$ and $\beta$ with Eq.~\ref{e10}, where $\nabla_{\alpha}\lambda_{i,j} = cossim(x_i,x_j) $, $ \nabla_{\beta}\lambda_{i,j} = 1$ and $\nabla_{\lambda} E$ is obtained with an estimator (see Eq.~\ref{e11_0}$\sim$\ref{e12}).
\begin{equation}
\begin{aligned}
   & \hat{\alpha} = \alpha - \gamma_2 \frac{1}{N^2}\sum_{i=1}^{N}\sum_{j=1}^{N}(\nabla_{\alpha}\lambda_{i,j}\nabla_{\lambda_{i,j}} E) \\
	&	\hat{\beta} = \beta - \gamma_2 \frac{1}{N^2}\sum_{i=1}^{N}\sum_{j=1}^{N}(\nabla_{\beta}\lambda_{i,j}\nabla_{\lambda_{i,j}} E) \\
	\end{aligned}
	\label{e10} 
\end{equation}

According to~\cite{r27}, continuous distributions have a simulation property that samples can be drawn from them in both direct and indirect ways, and for the general case $x\sim p_{(x;\theta)}$, we can draw a sample $\hat{x}$ in an indirect way by firstly sampling $\overline{x}$ from a simple base distribution ${p_{(\epsilon)}}$, which is independent of the parameters $\theta$, and then transforming $\overline{x}$ to $\hat{x}$ through a sampling path ${sp(\epsilon;\theta)}$. Therefore, the expectation $E_{p(x;\theta)}[f(x)]$ can be converted to $E_{p(\epsilon)}[f(sp(\epsilon;\theta))]$ with Eq.~\ref{e11_0}. Furthermore, based on the \textit{Law of the Unconscious Statistician} (LOTUS)~\cite{r27}, a path-wise estimator for gradient ${\nabla_{\theta} E_{p(x;\theta)}[f(x)]}$ can be calculated with Eq.~\ref{e11}.

\begin{equation}
    E_{p(x;\theta)}[f(x)] = E_{p(\epsilon)}[f(sp(\epsilon;\theta))],
		\label{e11_0}
\end{equation}

\begin{equation}
	\begin{split}
	 \nabla_{\theta} E_{p(x;\theta)}[f(x)] &= \nabla_{\theta} \int p(\epsilon)f(sp(\epsilon;\theta))\mathrm{d}\epsilon\\
	 &=\int p(\epsilon)\nabla_{x} f(x)|_{x=sp(\epsilon;\theta)}\nabla_{\theta}sp(\epsilon;\theta)\mathrm{d}\epsilon\\
	 &=E_{p(x;\theta)}[\nabla_{x} f(x)\nabla_{\theta} x]
	\end{split}
		\label{e11} 
\end{equation}

However, in our case, the distribution $A\sim P$ is discontinuous because elements of $A$ are binarized, so we approximately estimate $\nabla_\lambda E$ with an inexact but smooth reparameterization of $A\sim P$ (see Eq.~\ref{e12}). 
Specifically, we employ the identity mapping $A=sp(\epsilon;\lambda)=1 - Po(0;\lambda)$ of straight-through estimators (STE)~\cite{r29}, and accordingly, get ${|\nabla_\lambda A| = |\nabla_\lambda Po(0;\lambda)| \approx I}$.
\begin{equation}
	\begin{aligned}
	 \nabla_\lambda E &= \nabla_\lambda E_{A\sim P}[\ell(f(\hat{X}), y)] \\
	&= E_{A\sim P}[\nabla_\lambda A \nabla_A \ell(f(g(X, W, A)), y)]\\
    &\approx \frac{1}{S}\sum_{s=1}^{S}\nabla_A \ell(f(g(X, W, A_s)), y)
	\end{aligned}
	\label{e12} 
\end{equation}

\section{Experiments}
In this section, we give details of our implementation and conduct extensive experiments to prove the effectiveness and robustness of our AGFN on datasets with uncertainties.
\subsection{Datasets and Implementation}
RAF-DB~\cite{r19} contains 12,271 training images and 3,068 test images collected from thousands of individuals. In our experiments, only images belonging to the 7 basic expressions (i.e., neutral, happiness, surprise, sadness, anger, disgust and fear) are used.

FERPlus~\cite{r20} is a large-scale dataset collected by Google search engine. It consists about 28,000 training images and 3,000 test images. Compared with RAF-DB, FERPlus includes an extra expression, i.e., contempt, resulting in 8 expression classes. In addition, since each sample in FERPlus is labeled by 10 annotators, we select label with the highest score as its ground truth. 

In our implementation, we embed the proposed AGFN into a naive FER baseline model, who employs ResNet-18 as its feature extractor and a fully-connected layer as its expression classifier. The projected emotional feature vectors are with a dimension of 512 and the batch size is set to 256. 
Moreover, we set the learning rates $\gamma_1$ and $\gamma_2$ (used in Eq.~\ref{e8} and ~\ref{e10}) to 0.01 and 0.001, respectively, and to speed the training procedure, we pre-train the baseline model for about 10 epochs before integrating the proposed AGFN module.

\subsection{Comparison with Existing FER Models}

In Table~\ref{Tab_1}, we compare the proposed AGFN-equipped network with existing FER works to demonstrate the effectiveness of our AGFN and the importance of dealing with data uncertainties in FER. Apparently, our network outperforms all of the listed methods on both FERPlus and RAF-DB datasets, and from the comparison with our baseline model, recognition accuracies are elevated significantly from 85.37\% to 91.84\% and from 85.89\% to 91.11\% on FERPlus and RAD-DB dataset, respectively by integrating the proposed AGFN module.

Recently, Transformer~\cite{r26} has largely fueled the performance of computer vision tasks, including FER. For example, FER-VT~\cite{r22} exploits grid-wise attention and visual Transformer to learn long-range inductive biases between different facial regions, and TransFER~\cite{r33} learns rich relation-aware local representation with Transformer. However, thought FER-VT and TransFER surpass other existing methods significantly in Table~\ref{Tab_1}, our AGFN-equipped network achieves better performance than them, demonstrating the effectiveness of our network. Moreover, we also conduct ablation study with methods employing GCN, e.g., GA-FER~\cite{r31} and FDRL~\cite{r32}, since GCN has been widely used in FER. Here, GA-FER~\cite{r31} builds graphs upon landmarks of single face images, while FDRL~\cite{r32} constructs graphs with latent features obtained by decomposing basic features of face images. On the other hand, GA-FER~\cite{r31} exploits a GCN equipped with complicated attention mechanism, which FDRL~\cite{r32} and our network use the naive GCN. Obviously, our network achieves better performance than both GA-FER~\cite{r31} and FDRL~\cite{r32}. In addition, we also compare our network with a model named Baseline-GCN, where GCN is integrated into the baseline model and samples with similarities greater than 0.5 are directly connected to build topological graphs. Apparently, our network obtains much higher accuracies than Baseline-GCN, indicating the superiority of our Poisson graph generator over the widely used threshold-based staircase function.

On the other hand, SCN~\cite{r01} also improves FER performance by tackling with the data uncertainty problem. It relabels mislabeled data by weighting and ranking samples in a mini-batch with a self-attention mechanism. In contrast, our AGFN protects FER models from data uncertainties by alleviating the excursive semantic and feature covariate shifting problem with associative relations of expressions. Obviously, our network surpasses SCN~\cite{r01} by 2.49\% and 2.91\% on FERPlus and RAF-DB, indicating that AGFN is with more advantages than SCN~\cite{r01} when dealing with the data uncertainty problem. 

\begin{table}[h]
\caption{\textbf{Comparison with existing works on RAF-DB and FERPlus datasets. Recognition accuracy is the metric.}}
\label{Tab_1}
\centering
\begin{tabular}{lll}
\toprule
 Method & FERPlus & RAF-DB \\ \midrule
SeNet~\cite{r21} & 88.8 &  -\\
RAN~\cite{r10} & 88.55 & 86.90 \\ 
DLP-CNN~\cite{r19} & -& 84.22\\
SCN~\cite{r01} & 89.35& 88.14\\ 
Baseline & 85.37 & 85.89\\
\hline
FER-VT~\cite{r22} & 90.04 & 88.26 \\
TransFER~\cite{r33} & 90.83 & 90.91\\
\hline
GA-FER~\cite{r31} & - & 87.52 \\
FDRL~\cite{r32} & - & 89.47 \\
Baseline-GCN & 88.8 & 88.95 \\
Our network & \textbf{91.84}& \textbf{91.11}\\ \bottomrule
\end{tabular}
\end{table}

\subsection{Performance of AGFN on Datasets with Serious Uncertainties}

With the growth in scale of training samples gathered from internet, the problem of data uncertainty is getting more and more severe. To further explore the effectiveness of AGFN on datasets with serious uncertainties, we conduct extra experiments on our synthetic datasets. Specifically, we randomly select 10\% or 20\% samples from FERPlus and RAF-DB datasets, and assign wrong labels to them. The comparison results are shown in Table~\ref{Tab_2}, where the proposed network is compared with SCN~\cite{r01} and other two state-of-the-art noise-tolerant methods, i.e., CurriculumNet~\cite{r23} and MetaCleaner~\cite{r24}. CurriculumNet~\cite{r23} handles massive amount of noisy labels and data imbalance on large-scale web images by leveraging curriculum learning, which measures and ranks the complexity of data in an unsupervised manner, while MetaCleaner~\cite{r24} learns to hallucinate a clean representation for an object category according to a small noisy subset from the same category.
From Table~\ref{Tab_2}, the recognition accuracy of our network is obviously higher than that of other three methods. Especially, from the comparison with SCN~\cite{r01}, our network achieves accuracy improvements of 2.75\% and 4.84\% on FERPlus and RAF-DB datasets with 10\% mislabeled samples, and the figures are 3.38\% and 4.52\% on datasets with 20\% mislabeled samples. Therefore, our AGFN-equipped network is with better effectiveness and robustness than other works when the data becomes more noisy.    

%

\begin{table}[h]
\caption{\textbf{Comparison results on synthetic datasets with noise ratios of 10\% and 20\%.}}
\label{Tab_2}
\centering
\begin{tabular}{c|cc|cc}
\hline
\multirow {2}{*}{Method} & \multicolumn{2}{c|}{FERPlus}    & \multicolumn{2}{c}{RAF-DB}      \\ \cline{2-5} 
                        & 10\%           & 20\%           & 10\%           & 20\%           \\ \hline
CurriculumNet           & -              & -              & 68.50          & 61.23          \\
MetaCleaner             & -              & -              & 68.45          & 61.35          \\
SCN                     & 84.28          & 83.17          & 82.18          & 80.10          \\
Our network             & \textbf{87.03} & \textbf{86.55} & \textbf{87.02} & \textbf{84.62} \\ \hline
\end{tabular}
\end{table}

\subsection{Performance on Datasets with Occlusion and Pose Variant}

In practice, occlusion and pose variation frequently happen and result in data uncertainties. Therefore, follow RAN~\cite{r10}, we also conduct additional experiments on Occlusion-FERPlus, Pose-FERPlus, Occlusion-RAF-DB and Pose-RAF-DB, which are generated from FERPlus and RAF-DB by Wang et al~\cite{r10}, to evaluate the performance of our network. The experimental results are listed in Table~\ref{Tab_4} and Table~\ref{Tab_5}. Here, RAN~\cite{r10} adaptively captures the importance of facial regions for occlusion and pose variant FER, and CVT~\cite{r25} translates facial images into sequences of visual words and performs expression recognition from a global perspective with convolutional visual Transformers. As shown in Table~\ref{Tab_4} and Table~\ref{Tab_5}, our network outperforms all of the listed methods on datasets with occlusion and achieves comparable performance with the latest Transformer-based works, i.e., CVT~\cite{r25} and FER-VT~\cite{r22}, on datasets with pose variation.
 
\begin{table}[h]
\caption{\textbf{Performance on datasets with occlusion.}}
\label{Tab_4}
\centering
\begin{tabular}{ccc}
\toprule
Method& Occlusion-FERPlus& Occlusion-RAF-DB \\
\midrule
RAN~\cite{r10}  & 83.63& 82.72\\
CVT~\cite{r25} & 84.79& 83.95\\
FER-VT~\cite{r22} & 85.24& 84.32\\
Our network & \textbf{85.95}& \textbf{86.53}\\
\bottomrule
\end{tabular}
\end{table}

\begin{table}[h]
\caption{\textbf{Performance on datasets with pose variation.}}
\label{Tab_5}
\centering
\begin{tabular}{ccc}
\toprule
Method& Pose-FERPlus & Pose-RAF-DB\\
\midrule
RAN~\cite{r10}  & 82.23& 85.20\\
CVT~\cite{r25} & 88.29& \textbf{88.35}\\
FER-VT~\cite{r22} & \textbf{88.56}& 86.08\\
Our network & 84.87& \textbf{88.35}\\
\bottomrule
\end{tabular}
\end{table}

\subsection{Parameter Sensitivity Analysis}

\begin{figure}[h]
\begin{center}
\includegraphics[width=9cm]{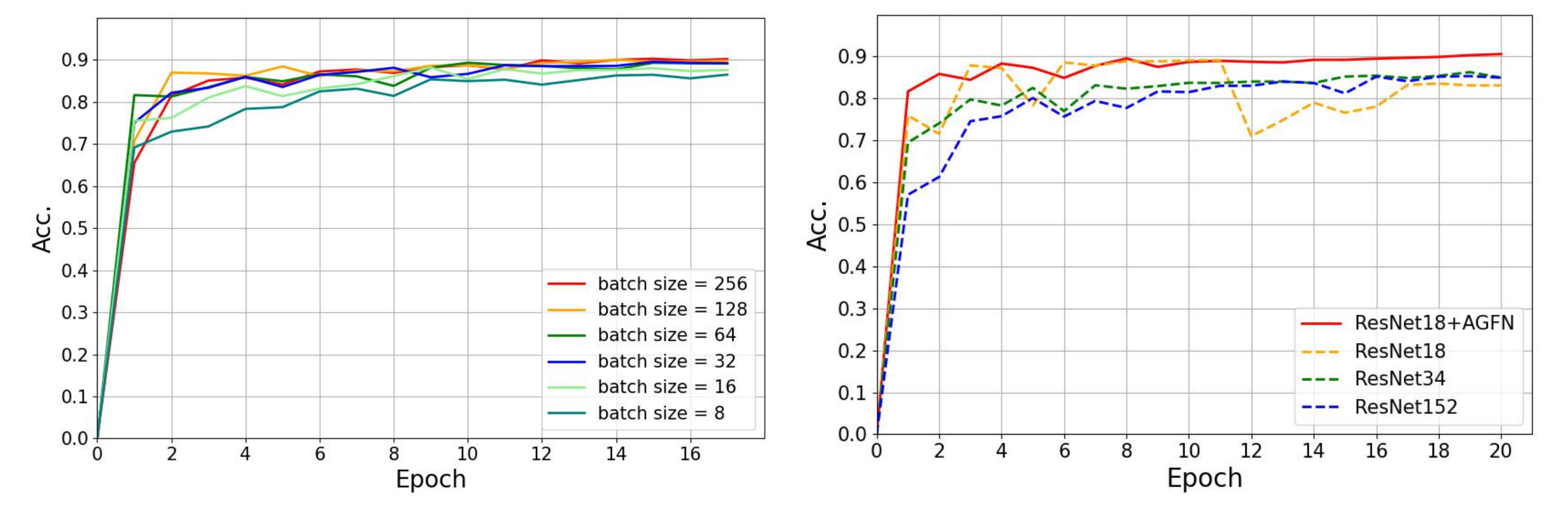}
\end{center}
\caption{Performance of our network with different batch size (left) and comparison of networks with AGFN and deeper backbones (right) on RAF-DB dataset. }
\label{fig:batchsize-backbone}
\end{figure}

In this part, we analyze the parameter sensitivity of our AGFN-equipped network in terms of batch size and backbone depth. Firstly, since our Poisson graph generator constructs topological graphs for samples within mini-batches, the setting of batch size is important. Therefore, as shown in Figure~\ref{fig:batchsize-backbone}, we evaluate the effect of different batch size settings to the performance of proposed network. As we can see, when the batch size is set to greater than 16, the performance is barely changed, and for batch size less than 16, the drop of performance is also in a reasonable range, so our proposed network is with strong robustness.  

\begin{table}[h]
\caption{\textbf{Comparison of networks with AGFN and different backbones on RAF-DB dataset.}}
\label{Tab_6}
\centering
\begin{tabular}{cc|cc}
\toprule
Method & Acc. & Method & Acc.\\
\midrule
ResNet18 & 85.89 & ResNet18+AGFN & 91.11\\
ResNet34 & 85.95 & ResNet34+AGFN & 90.38\\
ResNet152 & 86.02 & ResNet152+AGFN & 87.22\\
\bottomrule
\end{tabular}
\end{table}

\begin{figure*}[tp]
\begin{center}
\includegraphics[width=8cm]{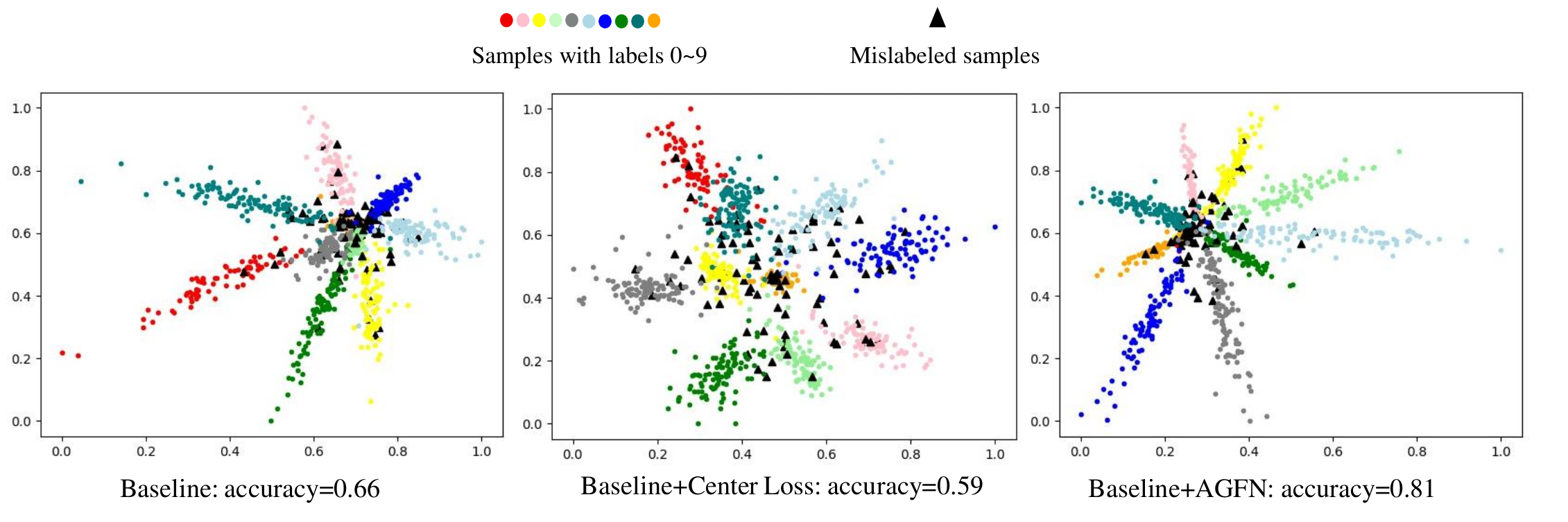}
\end{center}
\caption{Comparison results on MNIST.}
\label{fig:mnist}
\end{figure*}

\begin{figure*}[h]
\begin{center}
\includegraphics[width=12cm]{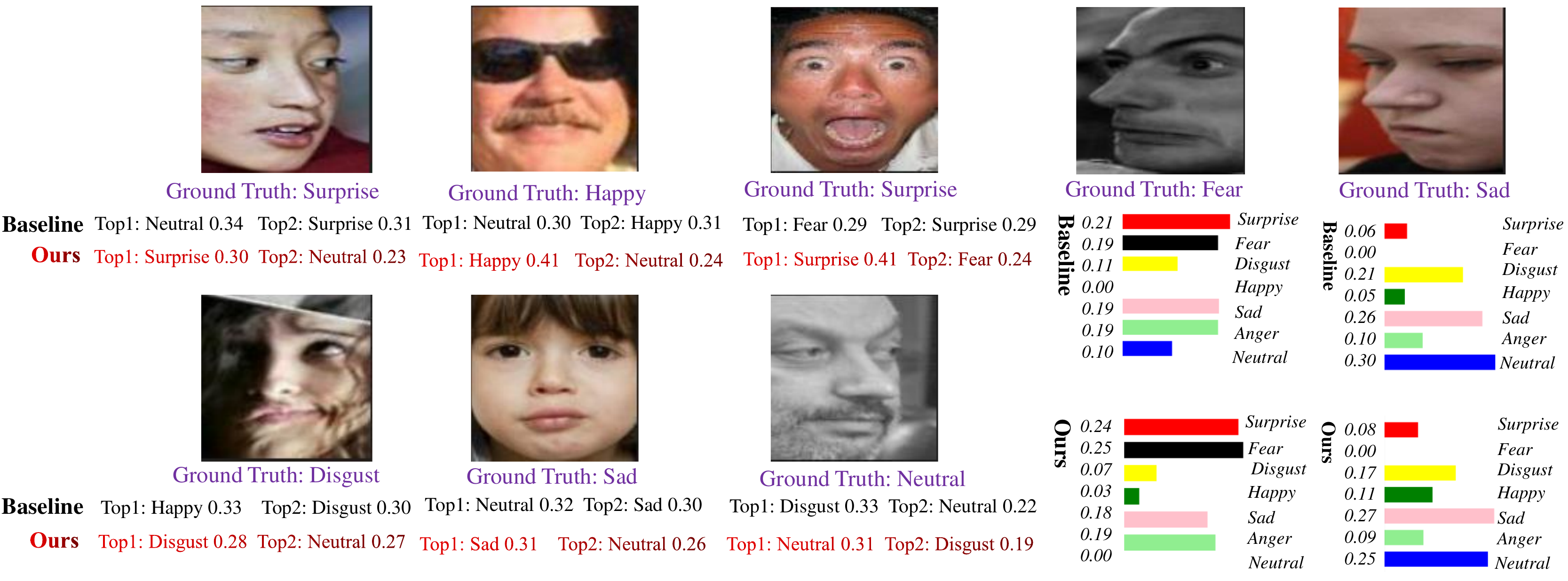}
\end{center}
\caption{Visualization results of samples with ambiguity.}
\label{fig:vis}
\end{figure*}

On the other hand, our AGFN module introduces additional parameters to the baseline network. It is a common sense that increasing the scale of deep networks is an effective way to enhance model's ability. Therefore, to prove that the performance improvement of our network is contributed by the strategy of utilizing associative relations of expressions rather than increasing model's complexity, we conduct extra experiments in Figure~\ref{fig:batchsize-backbone} and Table~\ref{Tab_6}, where performances of networks with proposed AGFN and different backbones are compared. Apparently, no matter what backbone is used, the performance can always be obviously improved by integrating proposed AGFN, especially,  when ResNet18 is exploited, the accuracy is elevated significantly from 85.89\% to 91.11\%.
Moreover, from the comparison of networks with ResNet18, ResNet34, ResNet152 and ResNet18+AGFN, we can conclude that integrating AGFN is more effective than simply increasing network's depth. Besides, from the first column of Table~\ref{Tab_6}, the accuracy is barely improved when increasing network's depth from 18 to 152, implying that for data with serious uncertainties, expanding the scale of networks does not work well. Additionally, we blame the performance degradation from ResNet152+AGFN to ResNet18+AGFN to the over-fitting problem, which is another reason why the performance is barely improved when changing ResNet18 to ResNet152 in the baseline model.  

\subsection{Generalizability of Proposed AGFN}

To evaluate the generalizability of proposed AGFN on dealing with data uncertainties, we conduct experiments on the well-known MNIST dataset, as shown in Figure~\ref{fig:mnist}. Here, we randomly select 10,000 samples from the training set of MNIST and assign wrong labels to 10\% of them. These samples form our training set and the original test set of MNIST is kept as our test set. From the distributions of samples in Figure~\ref{fig:mnist} we can see that our AGFN-equipped network is able to effectively decrease the intra-class variance and, meanwhile, increase the inter-class variance, resulting in a better recognition accuracy of 81\%, which is 15\% higher than that of our baseline model. 

On the other hand, Center Loss~\cite{r27} is widely used in various computer vision tasks. It also utilizes relations of samples to enhance the discriminative power of extracted features. However, different from our AGFN, Center Loss is a supervision signal who builds connections among samples according to their labels, while our AGFN constructs topological graphs in an unsupervised way by utilizing samples' similarities. On the other hand, Center Loss simultaneously learns a center for deep features of each classes and penalizes distances between deep features and their corresponding class centers. In contrast, our AGFN conveys semantic information among samples with GCN to normalize the distribution of features. Therefore, mislabeled data would affect the learning of class centers in Center Loss, resulting in performance degradation, while our AGFN is able to protect FER models from mislabeled data. Comparison results of their effects to feature distribution are present in Figure~\ref{fig:mnist}.

\subsection{Visualization results}
To intuitively understand the effectiveness of our AGFN-equipped network, we present the visualization results for samples with ambiguity in Figure~\ref{fig:vis}. From samples in the first three columns, the baseline model usually assigns similar scores to its top-2 predictions and fails to generate correct predictions. In contrast, our network not only generates correct predictions but also predicts top-2 scores with relatively larger distance. Moreover, from the fourth sample in the first row, the baseline model fails to distinguish the `Fear' expression from `Surprise', `Sad' and `Anger'. It assigns the ground truth label `Fear' a score of 0.19, which is the same as that of `Sad' and `Anger', and lower than that of `Surprise'. In contrast, our network predicts the expression as `Fear' with a score of 0.25, which is obviously higher than that of 'Sad' and `Anger'.        

\section{Conclusion}
This paper proposes to utilize the associative relations of expressions to tackle the excursive semantic and feature covariate shifting problem caused by data uncertainties in FER. It presents an effective feature normalization method named AGFN, who exploits a Poisson graph generator to dynamically and adaptively construct topological graphs for samples in each mini-batches, and employs GCN to convey semantic information among samples. Additionally, to jointly optimize parameters involved in neural networks and the sampling process, a coordinate decent strategy is designed. Extensive experiments demonstrate the effectiveness of proposed AGFN and the importance of addressing the data uncertainty problem. The proposed network not only outperforms existing works on the benchmark datasets FERPlus and RAF-DB but also surpasses state-of-the-art works by 3.38\% and 4.52\% when the percentage of mislabeled data significantly increases (i.e., to 20\%).    

\clearpage

{\small
\bibliographystyle{ieee_fullname}
\bibliography{egbib}
}

\end{document}